\pdfoutput=1

\documentclass[11pt]{article}

\usepackage[]{acl}
\usepackage[normalem]{ulem}
\useunder{\uline}{\ul}{}
\usepackage{times}
\usepackage{latexsym}
\usepackage{subfigure}
\usepackage[T1]{fontenc}

\usepackage[utf8]{inputenc}

\usepackage{microtype}

\usepackage{inconsolata}

\usepackage{graphicx} %
\usepackage{natbib}  %
\usepackage{caption} %
\usepackage{algorithm}
\usepackage{listings}
\usepackage{algorithmic}
\usepackage{amsmath}
\usepackage{booktabs}
\usepackage{multirow}
\usepackage[outdir=./]{epstopdf}
\usepackage{enumitem}
\usepackage{caption}
\usepackage{subcaption}
\usepackage{subfloat}
\usepackage{newfloat}
\usepackage{graphicx}
\usepackage{svg} %
\usepackage[normalem]{ulem}
\usepackage{framed}
\usepackage{mdframed}
\usepackage{xcolor}
\usepackage{lipsum}
\usepackage{float}
\usepackage{hyperref}
\usepackage[utf8]{inputenc}
\usepackage{forest}
\definecolor{shadecolor}{gray}{0.9}

\definecolor{verylightgray}{rgb}{0.96,0.96,0.96}

\newenvironment{styleditemize}
{\begin{mdframed}[backgroundcolor=verylightgray,
    skipabove=0.5em,
    skipbelow=0.5em,
    innertopmargin=0.5em,
    innerbottommargin=0.5em,
    innerleftmargin=0.5em,
    innerrightmargin=0.5em,
    roundcorner=4pt]
\begin{itemize}[topsep=0cm, itemsep=0cm, leftmargin=*]\small}
{\end{itemize}\end{mdframed}}
\newenvironment{styledenumerate}
{\begin{mdframed}[backgroundcolor=verylightgray,
    skipabove=0.5em,
    skipbelow=0.5em,
    innertopmargin=0.5em,
    innerbottommargin=0.5em,
    innerleftmargin=0.5em,
    innerrightmargin=0.5em,
    roundcorner=4pt]
\begin{enumerate}[topsep=0cm, itemsep=0cm, leftmargin=0.8cm]}
{\end{enumerate}\end{mdframed}}

\newlist{todolist}{itemize}{2}
\setlist[todolist]{label=$\square$}
\usepackage{pifont}

\usepackage{array}
\newcolumntype{L}[1]{>{\raggedright\let\newline\\\arraybackslash\hspace{0pt}}m{#1}}
\newcolumntype{C}[1]{>{\centering\let\newline  \\\arraybackslash\hspace{0pt}}m{#1}}
\newcolumntype{R}[1]{>{\raggedleft\let\newline \\\arraybackslash\hspace{0pt}}m{#1}}

\AtBeginDocument{%
  \providecommand\BibTeX{{%
    \normalfont B\kern-0.5em{\scshape i\kern-0.25em b}\kern-0.8em\TeX}}
}

\title{Open-Set Living Need Prediction with Large Language Models}
\author{Xiaochong Lan$^{1,2}$ \ \ \ \ Jie Feng$^1$\thanks{Corresponding author.} \ \ \ \ Yizhou Sun$^1$ \ \ \ \ Chen Gao$^1$ \\ \textbf{Jiahuan Lei}$^2$ \ \ \ \ \textbf{Xinlei Shi}$^2$ \ \ \ \  \textbf{Hengliang Luo}$^2$ \ \ \ \ \textbf{Yong Li}$^1$\footnotemark[1]\\
$^1$Department of Electronic Engineering, BNRist, Tsinghua University \ \ $^2$Meituan\\
\small \texttt{lanxc22@mails.tsinghua.edu.cn} \ \ \ \texttt{fengjie@tsinghua.edu.cn} \ \ \ \texttt{liyong07@tsinghua.edu.cn}
}
\begin{document}
\maketitle
\begin{abstract}
Living needs are the needs people generate in their daily lives for survival and well-being. On life service platforms like Meituan, user purchases are driven by living needs, making accurate living need predictions crucial for personalized service recommendations. Traditional approaches treat this prediction as a closed-set classification problem, severely limiting their ability to capture the diversity and complexity of living needs. In this work, we redefine living need prediction as an open-set classification problem and propose PIGEON, a novel system leveraging large language models (LLMs) for unrestricted need prediction. PIGEON first employs a behavior-aware record retriever to help LLMs understand user preferences, then incorporates Maslow's hierarchy of needs to align predictions with human living needs. For evaluation and application, we design a recall module based on a fine-tuned text embedding model that links flexible need descriptions to appropriate life services. Extensive experiments on real-world datasets demonstrate that PIGEON significantly outperforms closed-set approaches on need-based life service recall by an average of 19.37\%. Human evaluation validates the reasonableness and specificity of our predictions. Additionally, we employ instruction tuning to enable smaller LLMs to achieve competitive performance, supporting practical deployment.
\end{abstract}

\section{Introduction}
\label{sec::intro}
Living needs refer to the needs people generate in their daily lives for survival and well-being, from basic needs like food and accommodation to higher-level needs like socializing and hobbies. Living needs play a central role in life service platforms like Meituan\footnote{~\href{https://www.meituan.com}{meituan.com}}, which serves as a bridge between millions of life service providers and nearly a billion users~\cite{chen2022mixed,liu2022modeling,liu2025mrgrp,liang2024meituan}. On such platforms, user purchases are fundamentally driven by their underlying living needs - for example, when users order food delivery, it is typically motivated by their need for food. Therefore, accurate prediction of users' living needs is crucial for service recommendation, as it enables platforms to suggest services that directly address these needs, thereby improving recommendation accuracy and user experience.
\begin{figure}[t]
\centering
\includegraphics[width=6cm]{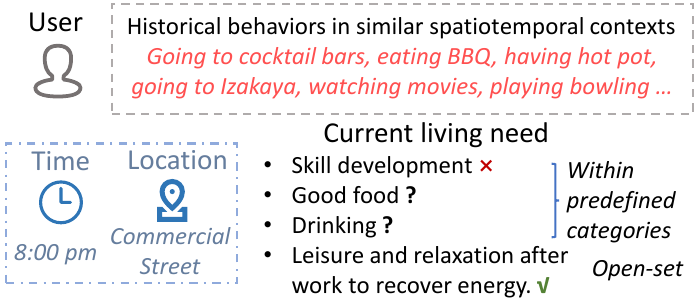}
\vspace{-0.2cm}
\caption{We define living need prediction as an open-set classification problem with unlimited classes to handle the complexity of living needs.}
\vspace{-0.4cm}
\label{fig:case}
\end{figure}

Living need prediction aims to predict a user's needs based on their spatiotemporal context. Previous studies have treated this as a \textbf{closed-set} classification problem, where models select from limited predefined need categories~\cite{ping2021user,lan2023neon}. However, this closed-set formulation has inherent limitations. First, users' needs are highly \textit{diverse}, making it impossible to list all possibilities. Additionally, users' needs may be \textit{vague}, such as seeking a relaxing atmosphere rather than specific venue types like restaurants or bars. Furthermore, users' needs can be \textit{composite}, such as wanting a filling while also healthy meal. 
Obviously, predicting living needs within a fixed set of categories cannot handle these complexities.

In this work, we reformulate living need prediction as an open-set classification problem with unlimited classes, moving beyond the constraints of predefined categories~\cite{zhang2021deep,bao2022opental}. Following insights that generative models show promise for open-set classification~\cite{mundt2019open} and building on recent advances~\cite{zareian2021open,pham2024lp}, we leverage large language models (LLMs) for this task. LLMs can generate unrestricted textual descriptions of living needs, naturally addressing the limitations of closed-set methods. Their extensive common sense knowledge enables deeper understanding of complex human needs. Through this approach, we aim to achieve truly open-set need prediction capable of identifying an unlimited range of living needs.

However, using LLMs for open-set living need prediction introduces additional challenges:
\vspace{-0.3cm}
\begin{itemize}[leftmargin=*]
\setlength{\itemsep}{0pt}
\setlength{\parsep}{0pt}
\setlength{\parskip}{0pt}
\item First, LLMs lack understanding of individual user preference. While LLMs hold extensive general knowledge, they lack understanding of specific user preferences. To bridge this gap, we must provide LLMs with relevant user behavior data. However, achieving optimal prediction accuracy requires careful selection and providing an appropriate number of highly relevant historical records~\cite{lin2024rella}, making the design of effective retrieval methods crucial.
\item Second, LLMs lack explicit knowledge of structured human living needs. While LLMs naturally support flexible need predictions, they sometimes produce descriptions that deviate from \textbf{needs}, like overly specific actions (e.g., \textit{``eat a KFC burger''}). Aligning predictions with a structured living needs framework is challenging.
\item Third, applying flexible need predictions in downstream tasks is difficult. The most typical application of need prediction is to recall life services that can fulfill a need. However, mapping a vast range of need types to the appropriate services is highly complex, making it challenging to model these associations effectively.
\end{itemize}
\vspace{-0.3cm}
To tackle these challenges, we introduce our system, PIGEON (short for o\textbf{P}en set l\textbf{I}vin\textbf{G} n\textbf{E}eds predicti\textbf{ON}). For the first challenge, unlike traditional text embedding-based retrieval methods, we design a historical interactions-based encoder to help retrieve the most relevant historical information for the LLM to understand individual preferences. To address the second challenge, we guide the LLM to use Maslow's theory to organize platform services, creating a structured framework of users' living needs. The LLM then uses this structure to refine its initial predictions. For the third challenge, we develop a fine-tuning approach for a text embedding model, creating a recall module capable of understanding complex relationships between flexible living needs and vast life services. Additionally, to make our method deployable on platforms, we design an instruction tuning approach that enables smaller LLMs to achieve competitive performance.

We compare our method with closed-set classification approaches and LLM-based baselines on recalling appropriate life services, and the experimental results demonstrate the effectiveness of our approach. Human evaluation confirms that our method generates predictions with superior reasonableness and specificity compared to closed-set methods. Cases illustrate that our method can accurately and comprehensively predict users' needs, handling vague and composite living needs. Ablation studies further validate the contributions of each component in our design. 

In summary, the main contributions of our work are as follows:
\vspace{-0.3cm}
\begin{itemize}[leftmargin=*]
\setlength{\itemsep}{0pt}
\setlength{\parsep}{0pt}
\setlength{\parskip}{0pt}
\item To the best of our knowledge, we are the first to explore the open-set living need prediction problem, which is a significant yet under-addressed challenge on life services platforms.
\item We propose PIGEON, an LLM-based open-set living need prediction system, which achieves state-of-the-art performance.
\item We design a recall approach for open-set needs and a fine-tuning method that enables smaller LLMs to achieve competitive performance, supporting the practical deployment of our method in online environments.
\end{itemize}
\vspace{-0.5cm}
\begin{figure*}[t]
\centering
\includegraphics[width=15cm]{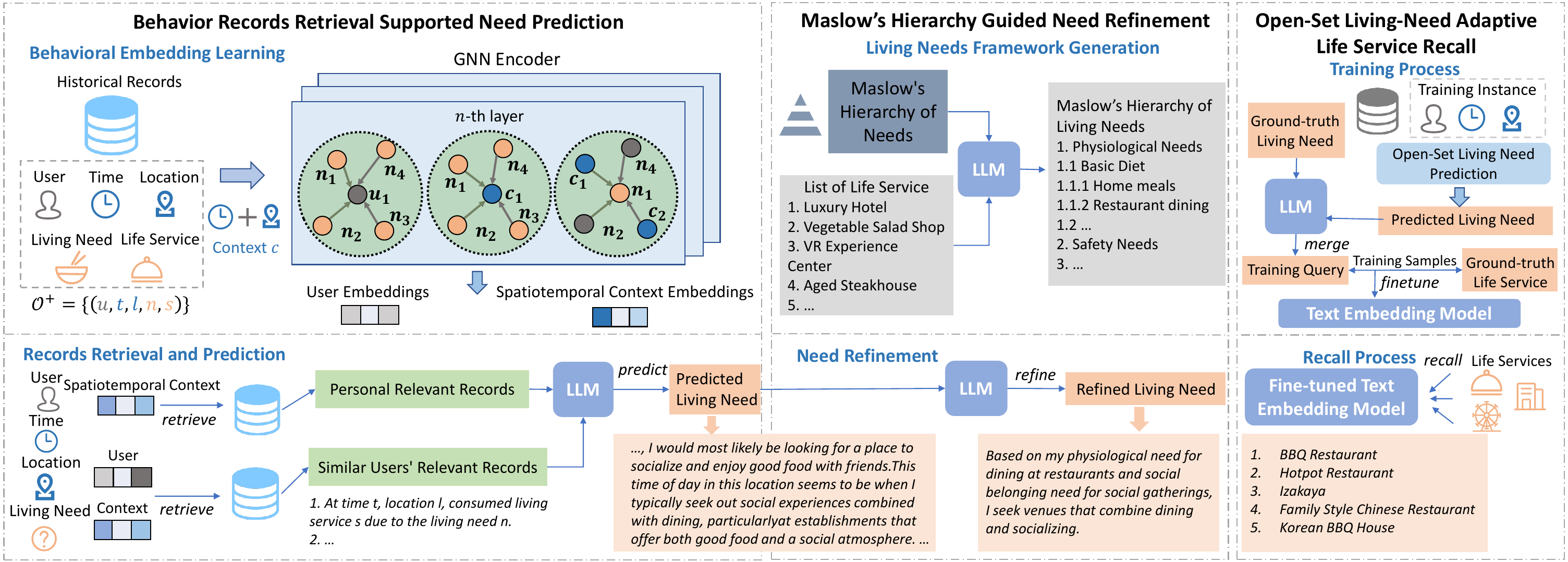}
\vspace{-0.3cm}
\caption{Architecture of our PIGEON system for open-set living need prediction and supporting recall module. First, a GNN encoder learns behavioral embeddings for users and spatiotemporal contexts, enabling retrieval of relevant historical records for LLM-based need prediction. To refine these predictions, the LLM incorporates Maslow's theory with platform services, creating a framework that aligns initial predictions with human needs. Additionally, a text embedding model is fine-tuned to link flexible need descriptions to relevant life services. To further support online deployment, the initial prompts to LLMs and their corresponding ground-truth-corrected predictions form instruction tuning pairs, which are used to fine-tune smaller LLMs.}
\vspace{-0.5cm}
\label{fig:main}
\end{figure*}
\section{Problem Formulation}
Living need prediction aims to predict a user’s living need at a given time and location. This prediction enhances user understanding and supports applications like life service recall. The ground-truth labels for this prediction task are derived from historical platform records, where user reviews sometimes reveal the living needs which drive the consumption. For example, a user might write, \textit{``I came here for a quick bite and found it quite good.''} Here, the user’s need can be identified as \textit{``a quick bite.''} Extracting and clustering such need descriptions, then refining them with expert input, allows us to establish a set of need categories. Traditionally, this set remains limited to enable living need classification. Formally, each historical consumption instance is represented by a five-tuple $(u,t,l,n,s)$, where $u$ is the user, $t$ the time, $l$ the location\footnote{
In this work, location is in the form of functional zones.}, $n$ the need type, and $s$ the consumed service. Thus, our problem is formulated as follows: 
\begin{itemize}[leftmargin=*] 
\vspace{-0.2cm} 
\item \textbf{Input}: $\mathcal{O}^{+}=\{(u, t, l, n, s) \mid u\in U, t\in T, l\in L, n\in N, s\in S\}$, where $U, T, L, N$, and $S$ denote the sets of users, time points, locations, need types, and life services, respectively.
\vspace{-0.2cm} 
\item \textbf{Output}: A model that, given $u$, $t$, and $l$, can infer the user’s type of living need as an unrestricted need description, covering limitless types. 
\end{itemize}

\section{Methods}\label{sec::method}
In this section, we provide a detailed description of each component of our method. Figure~\ref{fig:main} illustrates our proposed open-set living need prediction system along with the recall method that supports open-set need descriptions. The specific design of each component in our method is explained below.
\subsection{Behavior Records Retrieval Supported Living Need Prediction}
We aim to achieve open-set living need prediction with LLMs. Although LLMs possess extensive common sense, they lack specific knowledge of individual users. While explicit user profiles could provide such knowledge, they are difficult to obtain as they rely on users' active input. Instead, we leverage behavioral data that is naturally generated through user interactions. We augment LLM prediction with in-context learning from retrieved historical records~\cite{luo2024context,wang2024whole}, to help the LLM obtain knowledge about the specific user. Instead of traditional text embedding-based retrieval which only captures general text-level similarity, we propose behavioral embeddings learned from users' consumption histories. These embeddings directly encode individual preferences and contextual effects, enabling more precise retrieval of records for personalized predictions.

\paragraph{Behavioral Embedding Learning}
To encode the multiple relationships among users, spatiotemporal contexts, and needs, we use graph neural network (GNN) for embedding learning. First, we construct the graph. There are three types of nodes in the graph: user nodes $u$, spatiotemporal context nodes $c$, and living need nodes $n$. Since time and location jointly influence users' living needs, we combine them into unified spatiotemporal context nodes $c$. The graph has two types of weighted edges: user-need edges $(u,n)$ and context-need edges $(c,n)$. For each consumption record where user $u$ generates need $n$ in context $c$, we increment the weights of both $(u,n)$ and $(c,n)$ by 1.

Next, we describe the representation learning method on the graph. 
We first randomly initialize embeddings $\mathbf{e}^{(0)}_u$, $\mathbf{e}^{(0)}_c$, and $\mathbf{e}^{(0)}_n$ for users, contexts, and needs respectively. 
For all kinds of nodes, the propagation on the graph is as follows:
$$
\mathbf{e}_i^{(k+1)}=\sum_{j \in \mathcal{N}(i)} \frac{w_{i j}}{D_i \cdot D_j} \mathbf{e}_j^{(k)},
$$
where $\mathbf{e}_i^{(k)}$ is node $i$'s embedding at layer $k, w_{i j}$ is the weight of edge $(i,j)$, and $\mathcal{N}(i)$ denotes the set of neighboring nodes of $i$.  $D_i=\sqrt{\sum_{j^{\prime} \in \mathcal{N}(i)} w_{i j^{\prime}}}$ and $D_j=\sqrt{\sum_{i^{\prime} \in \mathcal{N}(j)} w_{i^{\prime} j}}$ are normalization factors. After $K$ propagation layers, we obtain the final embedding through layer combination and normalization:
$$
\mathbf{e}_i=\frac{\sum_{k=0}^K \mathbf{e}_i^{(k)}}{\left\|\sum_{k=0}^K \mathbf{e}_i^{(k)}\right\|}
$$
To train the GNN encoder, we construct a Bayesian Personalized Ranking (BPR) loss:
$$
\mathcal{L}_{\mathrm{BPR}}=-\sum_{\left(u, c, n, n^{\prime}\right) \in \mathcal{D}} \ln \sigma\left(\mathbf{e}_{u c}^{\top} \mathbf{e}_n-\mathbf{e}_{u c}^{\top} \mathbf{e}_{n^{\prime}}\right),
$$
where $\sigma$ is the sigmoid function, $\mathbf{e}_{u c}=\mathbf{e}_u+\mathbf{e}_c$, and $\mathcal{D}$ represents the set of training instances $\left(u, c, n, n^{\prime}\right)$, with $n$ as the positive need generated by user $u$ in context $c$ and $n^{\prime}$ a randomly sampled negative need. Finally, after supervised training on historical records, we obtain embeddings for users and spatiotemporal contexts that effectively capture user preferences and contextual influences.
\paragraph{Records Retrieval and Prediction} 
Next, we utilize an LLM to predict the living need of a user in a given spatiotemporal context, based on retrieved most relevant behavioral records. While a user's own historical records are essential for personalization, relying solely on individual history has limitations: users may have unexplored needs that they haven't yet acted upon, or they may be in novel situations where relevant personal records are limited. To address these limitations, we complement individual records with historical data from users who exhibit similar behavioral patterns.

For user $u$ in context $c$, we first retrieve personal relevant records. From the user's entire historical record, we select the $K_p$ historical instances with context embeddings $\mathbf{e}_{c,i}$ most similar to the current context embedding $\mathbf{e}_c$. For the relevant historical records of similar users, we compute $\mathbf{e}_{u c}=\mathbf{e}_u+\mathbf{e}_c$ and retrieve $K_s$ historical records with embeddings $\mathbf{e}_{u c,i}$ most similar to the current $\mathbf{e}_{u c}$, from all history records. The similarity is measured using cosine similarity, for example,
$\operatorname{sim}(\mathbf{e}_{c,i}, \mathbf{e}_c)=\frac{\mathbf{e}_{c,i}^{\top} \mathbf{e}_c}{\|\mathbf{e}_{c,i}\|\|\mathbf{e}_c\|}
$.
Using the retrieved historical records, we then employ the LLM to predict the user's needs in the current spatiotemporal context. The specific prompt is as follows:
\begin{center}
\begin{minipage}{0.92\linewidth}
    \begin{shaded}
    \textit{You are a user on a life service platform. At [time] in [location], what kind of living need are you most likely to have?}
    
    \textit{1. Your following past behaviors are provided for reference: [Personal Relevant Records]}

    \textit{2. Behaviors of other users in similar contexts are: [Similar Users' Relevant Records]}

    \textit{Considering the current time, location, and preferences indicated by previous consumption behaviors, please infer and describe your potential living needs.}
    \end{shaded}
\end{minipage}
\end{center}
Here, the personal relevant records and similar users' relevant records are provided to the LLM in textual form, including information on time, location, type of living need, and the consumed life service. For example: \textit{``At 5 PM at home, you generated the living need for a quick meal and ordered a hamburger delivery service.''} Using this prompt as input, the LLM outputs a prediction of living needs expressed in an unrestricted text format.
\subsection{Maslow’s Hierarchy Guided Living Need Refinement}
Now, we have obtained a flexible text-based prediction of living needs. While the LLM allows for flexible predictions, some outputs may not align well with living \textbf{needs}. For instance, certain predictions may focus excessively on specific actions rather than genuine needs, such as \textit{``eating a McDonald's burger''}, or suggest needs unsupported by life service platforms, like \textit{``writing emails''}. Since living needs on service platforms naturally range from basic physiological needs to higher-level social and self-development needs, we leverage the widely-accepted Maslow's hierarchy~\cite{mcleod2007maslow} as a theoretical foundation to guide the refinement of these predictions. Specifically, we use the LLM to integrate Maslow's hierarchy with the platform's service offerings, creating a structured framework that helps align predictions with genuine human living needs while preserving their flexibility. In Section~\ref{sec:maslow}, we further explain the rationale for using Maslow's theory.

To construct this framework, we prompt the LLM with a list of available life services:
\begin{center}
\begin{minipage}{0.92\linewidth}
    \begin{shaded}
    \textit{[Life Services List]}
    
    \textit{These are life services on a platform that can meet human living needs. Based on this list, along with Maslow's hierarchy of needs, please generate a three-tiered framework of human living needs, ensuring that each need can be fulfilled by the listed services.}
    \end{shaded}
\end{minipage}
\end{center}
For large-scale service lists in the production environment that exceed LLM's token limit, we either categorize services first or process them in sequential batches. Using the generated framework, we then refine the initial predictions:

\begin{center}
\begin{minipage}{0.92\linewidth}
    \begin{shaded}
    \textit{You are a user on a life service platform. An inference about your current living need at [time] in [location] is [initial prediction]. }
    \end{shaded}
\end{minipage}
\end{center}
\begin{center}
\begin{minipage}{0.92\linewidth}
    \begin{shaded}
    \textit{Please use the following human living needs framework to further refine this inference, making it align with the framework’s scope. Your response should be concise and as informative as possible, around 20 words.}
    
    \textit{[Human living needs framework]}
    \end{shaded}
\end{minipage}
\end{center}
\subsection{Open-Set Living Need-Adaptive Life Service Recall}
In practical applications, predicted needs are typically used as one source in the recall stage to identify suitable life services as candidate recommendations. For closed-set methods, associations between needs and services rely on manually set rules. However, the flexibility of LLM-generated needs poses challenges for recall, as managing the complex mapping between a limitless range of needs and the services that fulfill them is nearly impossible. To address this, we design a recall approach based on a fine-tuned text embedding model, leveraging the commonsense knowledge embedded within language models to enhance recall accuracy.

In our recall method, need descriptions serve as queries to identify relevant services within the whole corpus of services. To fine-tune the text embedding model, we construct training samples from historical data by pairing actual needs with the ordered life services as ground truth. To adapt the model to flexible need descriptions as queries, we generate flexible living needs predictions for historical records and then use the LLM to refine these predictions based on the ground-truth closed-set living needs $n$. The specific prompt is:
\begin{center}
\begin{minipage}{0.92\linewidth}
    \begin{shaded}
    \textit{The ground-truth living need is [living need]; a flexible need description is: [predicted need]. Based on the ground-truth living need, please revise the flexible need description to maintain flexibility without compromising accuracy. Provide only the revised flexible need description, approximately 20 words.}
    \end{shaded}
\end{minipage}
\end{center}

Now we obtain flexible yet accurate need descriptions to serve as training queries for recall. Each refined need description $q$ is encoded as a vector embedding $f(q)$ using a pretrained transformer-based text embedding model $f(\cdot)$. The training process optimizes the model with a triplet loss:
$$
\mathcal{L}=\sum_{\left(q, s, s^{\prime}\right) \in \mathcal{D}} \max \left(0, \operatorname{sim}(q,s^{\prime})-\operatorname{sim}(q,s)+\alpha\right)
$$
where $\operatorname{sim}(q,s)=\frac{f(q)^{\top} f(s)}{\|f(q)\|\|f(s)\|}$ is the cosine similarity between the query $q$ and the positive life service $s$ (the actual consumed service), $s^{\prime}$ is a randomly selected negative service, and $\alpha$ is a margin. This loss encourages the model to embed services that can fulfill the need closer to the query than those that cannot fulfill the need, enhancing recall precision.
During inference, we use the fine-tuned model  $f_{\text{finetune}}(\cdot)$ to encode both queries and services, recalling services with highest similarities.

\subsection{Domain Adaptation for LLMs}
Practical deployment of our method faces computational and latency constraints, requiring high performance under limited computing resources and low latency for online deployment. In our current framework, LLM inference represents the primary computational and latency bottleneck. Therefore, we employ instruction tuning to enhance the capabilities of smaller LLMs that have lower computational requirements and latency. Our method naturally defines input-output pairs: the input being the initial prompt containing spatiotemporal context and user history, and the output being the refined training query that balances both accuracy and flexibility. After processing numerous prediction tasks with high-performance LLMs, we naturally accumulate these pairs as instruction tuning data. Using this constructed training data, we fine-tune smaller LLMs. Notably, we adopt an end-to-end training approach, aiming to train LLMs that can achieve good results in a single inference step using data generated through the two-step framework. We implement full-parameter fine-tuning using accelerate~\cite{accelerate}.

\section{Experimental Setup}\label{sec::expset}
\subsection{Datasets}
We collect two large-scale datasets from the Meituan platform, each containing 120 days of user consumption records from a different city in China. These records include user details, timestamps, locations, actual life service consumption, and living needs extracted from reviews. In these datasets, time is categorized into 48 intervals of half-hour windows, and locations are grouped into 13 categories based on functionality (e.g., residential areas, commercial streets). Following~\cite{li2022automatically}, we partition the datasets by time, using the first 96 days for training, the next 12 days for validation, and the final 12 days for testing. Dataset statistics are provided in Table~\ref{tab:statistics}.
\begin{table}[]
\centering
\setlength{\tabcolsep}{0.6mm}
\begingroup\small
\resizebox{1\linewidth}{!}{
\begin{tabular}{ccccccc}
\hline
\textbf{Dataset}  & \textbf{\#Records} & \textbf{\#Users} & \textbf{\#Locs} & \textbf{\#Time} & \textbf{\#Needs} & \textbf{\#Services} \\ \hline
\textbf{Beijing}  & 6,942,123          & 37,226           & 14                   & 48              & 20               & 2026                     \\
\textbf{Shanghai} & 7,797,225          & 43,213           & 14                   & 48              & 20               & 2155                     \\ \hline
\end{tabular}}
\endgroup
\vspace{-0.2cm}
\caption{Statistics of our utilized datasets.}
\vspace{-0.5cm}
\label{tab:statistics}
\end{table}

\subsection{Implementation Details}
In our main experiments, we primarily use GPT-4o mini~\cite{openai2024gpt4o} as the LLM backbone for our method, chosen for its ease of use, relatively high performance, and cost-effectiveness. The cost is around \$1.05 for 10,000 times of predictions. To maximize reproducibility, we set the temperature parameter to 0. We also report performance with other LLMs as backbones. For the text embedding model used as the recall module, we adopt bge-base-v1.5~\cite{xiao2023c}. We implement the GNN encoder using the PyTorch~\cite{paszke2019pytorch} framework. We implement graph propagation efficiently using sparse matrix operations. Each reported result is the average of 5 runs. 
\footnote{Code is available at \url{https://github.com/tsinghua-fib-lab/PIGEON}}

\subsection{Evaluation Metric}

Evaluating flexible need predictions poses inherent measurement challenges due to their open-ended nature. We address this by assessing prediction quality through the downstream task of life service recall, where need predictions serve as queries to retrieve relevant services. For quantitative evaluation, we adopt two widely used metrics: Recall@K and NDCG@K. Recall@K is defined as the proportion of actually consumed services that appear in the top K recalled results. NDCG@K further emphasizes that ground-truth services should be ranked high in the recalled list. While real-world recall systems focus less on NDCG, we include it here to better evaluate our recall system's ability to identify services that precisely meet user needs.
\subsection{Comparison Methods}
For life service recall, we compare our method against two types of methods: traditional closed-set living need prediction methods and LLM-based open-set need prediction methods. For closed-set methods, we include industry-standard CTR models: XGBoost~\cite{chen2016xgboost}, DeepFM~\cite{guo2017deepfm}, DCN~\cite{wang2017deep}, XDeepFM~\cite{lian2018xdeepfm}, EulerNet~\cite{tian2023eulernet}, which take user, time, and location as input features. We also evaluate graph-based models: LightGCN~\cite{he2020lightgcn}, HAN~\cite{wang2019heterogeneous}, DisenHCN~\cite{li2022disenhcn} that capture complex relationships through graph incorporating user, time, location, need, and service nodes and propagation on the graph. To maximize the recall performance potential of closed-set need prediction methods, we utilize combinations of closed-set needs with time and locations as queries during both the training and prediction phases of the recall model.

For LLM-based open-set methods, we test zero-shot CoT~\cite{kojima2022large}, which predicts solely from current context without historical data. We also include two retrieval-enhanced approaches: ReLLa~\cite{lin2024rella} and LLMSRec-Syn~\cite{wang2024whole}. Both approaches leverage text embeddings for historical record retrieval, where ReLLa retrieves from individual user records while LLMSRec-Syn additionally incorporates records from users with similar behaviors.  For both LLM-based methods, the recall model's training and usage are consistent with our approach, differing only in flexible need generation. A more detailed description of comparison methods is in Section~\ref{sec:baselines}.

\section{Experimental Results}
We conducted extensive experiments to address the following research questions:
\begin{itemize}[leftmargin=*]
\setlength{\itemsep}{0pt}
\setlength{\parsep}{0pt}
\setlength{\parskip}{0pt}
\vspace{-0.2cm} 
\item \textbf{RQ1:} How is the performance of our open-set need prediction method on life service recall? 
\item \textbf{RQ2:} How is the quality of living need expressions generated by our method?
\item \textbf{RQ3:} Does each component of our method contribute to the final performance in service recall? 
\item \textbf{RQ4:} How does domain adaptation improve the performance of smaller open-source models?
\item \textbf{RQ5:} How does our model perform on different LLM backbones?
\item \textbf{RQ6:} How do hyperparameter settings impact the performance of our method? 
\end{itemize}
Due to space constraints, we present the experimental results for RQ6 in the Appendix.

\subsection{Recall Performance~(RQ1)}

\begin{table*}[ht]
\centering
\setlength{\tabcolsep}{0.6mm}

\caption{Comparison of PIGEON and baselines on the life service recall task. Bold and underlined refer to the best and 2nd best performance. * indicates that PIGEON outperforms the best baseline with $p < 0.05$ in the paired t-test.}
\vspace{-0.2cm}
\label{tab:main}
\begingroup
\resizebox{0.9\linewidth}{!}{
\begin{tabular}{cc|cccc|cccc}
\hline
\multirow{2}{*}{\textbf{Category}} &
  \multirow{2}{*}{\textbf{Model}} &
  \multicolumn{4}{c|}{\textbf{Beijing}} &
  \multicolumn{4}{c}{\textbf{Shanghai}} \\
                                        &       & Recall@10     & Recall@20    & NDCG@10     & NDCG@20     & Recall@10     & Recall@20    & NDCG@10     & NDCG@20     \\ \hline
\multirow{8}{*}{Closed-Set} & XGBoost  & 0.03106       & 0.05878       & 0.01354  & 0.02032   & 0.02164       & 0.04774       & 0.00929  &   0.01600 \\
                                       & DeepFM   & 0.03378       & 0.06486       & 0.01546  &  0.02319   & 0.06047       & 0.09230       & 0.02903 &  0.03710   \\
                                       & DCN      & 0.03176       & 0.06351       & 0.01558  & 0.02356    & 0.06122       & 0.09377       & 0.03018  &  0.03862  \\
                                       & XDeepFM  & 0.03331       & 0.06552       & 0.01623   &0.02410    & 0.06203       & 0.09448       & 0.03081   &  0.03971  \\
                                       & EulerNet & 0.03497       & 0.06871       & 0.01687   & 0.02485   & 0.06343       & 0.09615       & 0.03109   &  0.03927 \\
                                       & LightGCN & 0.05510       & 0.09410       & 0.02624   & 0.03594   & 0.04375       & 0.07701       &0.01939  &  0.02785  \\
                                       & HAN      & 0.08851       & 0.12973       & 0.04532    & 0.05574  & 0.06111       & 0.09612       & 0.02745   &  0.03620  \\
                                       & DisenHCN & {\ul 0.09244} & {\ul0.13347} & {\ul 0.04820} &{\ul 0.05874} &{\ul 0.07268} & {\ul 0.12335} & 0.03474&   0.04377\\ \hline
\multirow{3}{*}{Open-Set} &
  Zero-shot CoT &
  0.03446 &
  0.05743 &
  0.01748 &
  0.02330 &
  0.02355 &
  0.04710 &
  0.01089&0.01680 \\                                       & ReLLa  & 0.05878       & 0.10405       & 0.03029  & 0.04150   & 0.05347       & 0.09612       & 0.02933    &0.03987   \\
 & LLMSREC-Syn  & 0.08044       &0.12705       & 0.04421  & 0.06012    & 0.06317      & 0.10025       & {\ul 0.03738} &  {\ul0.04792}   \\
 &
  PIGEON &
  \textbf{0.10405}$^*$ &
  \textbf{0.16622}$^*$ &
  \textbf{0.05124}$^*$ &
  \textbf{0.06681}$^*$ &
  \textbf{0.10503}$^*$ &
  \textbf{0.14895}$^*$ &
  \textbf{0.04743}$^*$ &
  \textbf{0.05845}$^*$ \\ \hline
\end{tabular}}
\endgroup
\end{table*}

We measure the performance of PIGEON and baselines in the life service recall task on two datasets, as shown in Table~\ref{tab:main}. Key findings are as follows:
\begin{itemize}[leftmargin=*]
\setlength{\itemsep}{0pt}
\setlength{\parsep}{0pt}
\setlength{\parskip}{0pt}
\item \textbf{Our method achieves top performance across all metrics on both datasets.} 
On Beijing and Shanghai datasets, our method significantly outperforms the best baseline across all metrics, with relative improvements of 15.18\% and 23.55\%, respectively. This highlights that our flexible need predictions effectively boost life service recall, leading to more accurate recommendations.
\item \textbf{The approach for historical record retrieval is crucial.} 
Among open-set methods, zero-shot CoT shows the lowest performance, indicating that common sense alone is insufficient for need prediction and that historical data is crucial. ReLLa and LLMSREC-Syn, which use text embeddings to retrieve records, also underperform compared to our method, underscoring the effectiveness of our behavior-based representation learning.
\item \textbf{Graph-based approaches lead in closed-set methods.} 
Graph neural networks model complex relationships and cross-influences better than CTR models, which rely on feature interactions. This finding supports the suitability of GNN encoding in our approach.

\end{itemize}

\begin{figure}[]
\centering
\includegraphics[width=0.9\columnwidth]{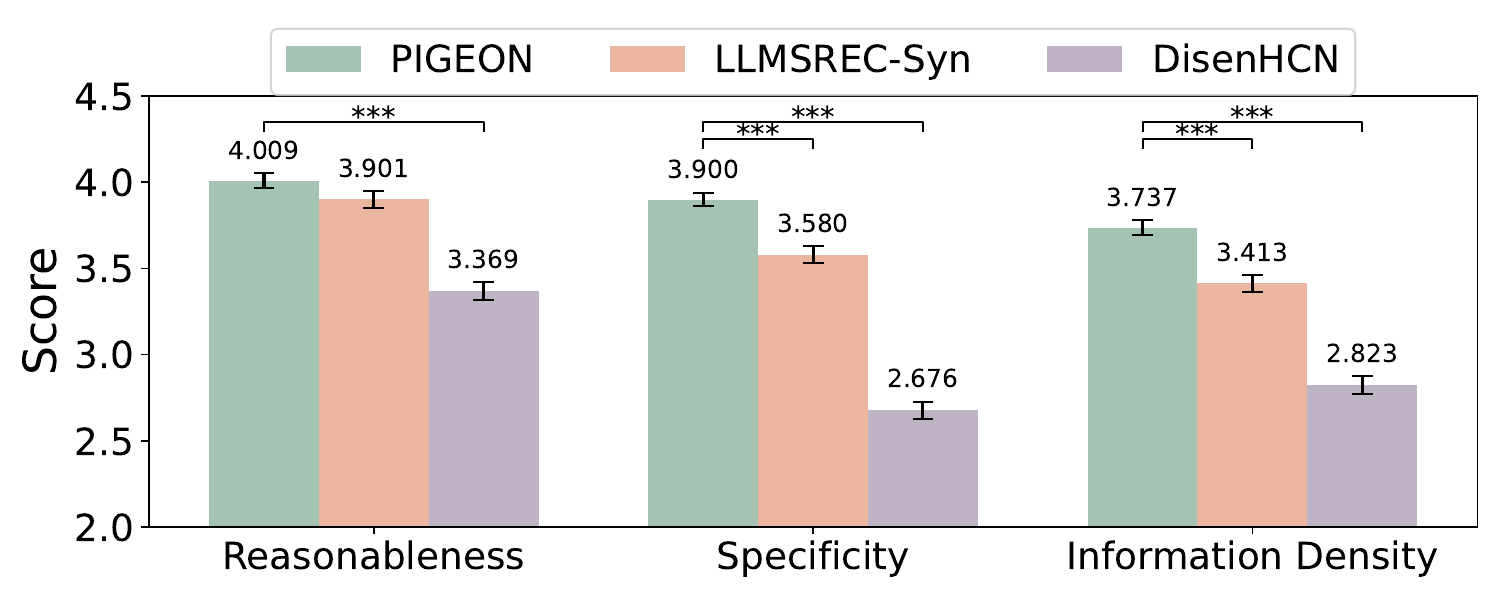}
\vspace{-0.2cm}
\caption{Human evaluation results. *** indicates $p<0.001$ in statistical significance test.} 
\vspace{-0.4cm}
\label{fig:humanplot}
\end{figure}

\subsection{Quality of Generated Prediction~(RQ2)}
To assess the quality of predictions generated by PIGEON, we conduct a human evaluation with 116 domain experts, who rate predictions from PIGEON, LLMSREC-Syn, and DisenHCN on reasonableness, specificity, and information density. For fair comparison, LLM outputs are constrained to approximately 10 words. Details of the questionnaire design are in Appendix Section~\ref{sec:human}.

Results of human evaluation are shown in Figure~\ref{fig:humanplot}. Statistical analysis shows that PIGEON generates high-quality predictions across all dimensions. While PIGEON and LLMSREC-Syn achieve comparable rationality scores (4.009 vs.\ 3.901, $p=0.103$), PIGEON significantly outperforms in both specificity (3.900 vs.\ 3.580, $t=5.107$, $p<0.001$) and information density (3.737 vs.\ 3.413, $t=5.062$, $p<0.001$). Compared to DisenHCN, PIGEON shows substantial improvements across all metrics ($p<0.001$), with particularly large gains in specificity ($t=18.852$) and information density ($t=13.737$).

Notably, LLM-based methods provide richer and more comprehensive need expressions that participants find more specific and actionable, while maintaining high information density despite longer descriptions. In contrast, closed-set predictions, constrained by predefined categories, often appear too generic for specific contexts. 

\label{sec:cases}
\begin{figure}[t]
\centering
\includegraphics[width=7.5cm]{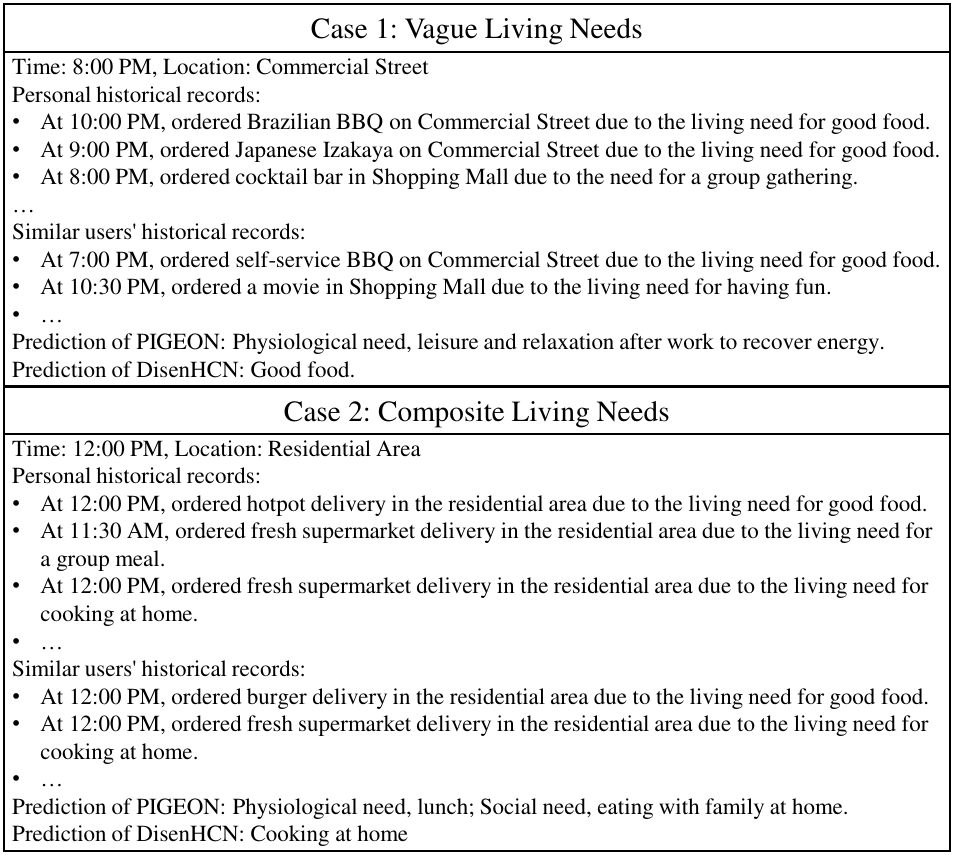}
\vspace{-0.2cm}
\caption{Cases show that our model can generate flexible living need descriptions covering vague and composite needs, while closed-set classification models cannot.}
\label{fig:case}
\vspace{-0.4cm}
\end{figure}

What's more, case studies in Figure~\ref{fig:case} demonstrate how our LLM-based approach addresses closed-set methods' limitations in capturing diverse, vague, and composite human needs.

In the first case, users ordered Brazilian barbecue, Japanese izakayas, and cocktail bars around 8 PM in commercial street. While closed-set methods categorize this as \textit{``good food''}, PIGEON recognizes the after-work context and infers a vague need: \textit{``relaxation and entertainment after work, to unwind and recharge''}.

In the second case, a user ordered food delivery and fresh produce around 12 PM in a residential area with gathering-related comments. PIGEON infers a composite need combining physiological (eating lunch) and social (family gathering) aspects, while the closed-set model DisenHCN only predicts a single need like ``cooking at home''.

These cases show our method effectively handles vague and composite needs, demonstrating clear advantages over closed-set classification methods.

\subsection{Ablation Study~(RQ3)}
To assess each component's impact on recall performance, we individually remove parts of PIGEON and present results in Table~\ref{tab:ablation} and~\ref{tab:ablation_shanghai}. The key findings are as follows:
\begin{itemize}[leftmargin=*]
\setlength{\itemsep}{0pt}
\setlength{\parsep}{0pt}
\setlength{\parskip}{0pt}
\item \textbf{Behavior Records Retrieval.} 
Removing either personal or similar users' historical record retrieval leads to a marked drop in performance, with both types of records showing a similar degree of impact. What's more, omitting all historical records causes an even more substantial decrease, highlighting the necessity of using actual behavior data in addition to general LLM common sense.
\item \textbf{Maslow Hierarchy-Guided Need Refinement.} 
Omitting the refinement process and relying only on the LLM’s initial prediction substantially reduces performance, highlighting the importance of aligning predictions with human needs.
\item \textbf{Recall Model Fine-tuning.} 
In our fine-tuning design, to create training queries that are both flexible and accurately aligned with life services, we first generate flexible need descriptions, then refine them using closed-set ground-truth needs from historical data. Our ablation experiments show performance drops when using either only closed-set needs or unrefined flexible needs, while using the original text embedding model without fine-tuning performs even worse. These results validate both our query construction approach and the necessity of fine-tuning for effective service recall.
\end{itemize}

\begin{table}[]
\centering
\footnotesize
\caption{Results of ablation study on Beijing dataset.}
\vspace{-0.2cm}
\label{tab:ablation}
\begingroup\small
\resizebox{1\linewidth}{!}{
\begin{tabular}{c|ccc}
\hline
\multirow{2}{*}{\textbf{Model}} & \multicolumn{2}{c}{\textbf{Beijing}} \\
                              & Recall@10 & Recall@20  \\ \hline
PIGEON                       & 0.10405 & 0.16622         \\
w/o history record          & 0.02770  & 0.05743        \\
w/o personal history        & 0.07838 & 0.13919    \\
w/o similar users' history     & 0.07808 & 0.13514      \\
w/o need refinement         & 0.09324  & 0.14932     \\
w/o recall model fine-tuning   & 0.07095 & 0.11486     \\ 
w/o flexible needs in queries  & 0.07500 & 0.12027       \\
w/o closed-set needs in queries & 0.09403    & 0.15176       \\
\hline
\end{tabular}}
\endgroup
\vspace{-0.1cm}
\end{table}

\begin{table}[]
\centering
\footnotesize
\caption{Results of ablation study on Shanghai dataset.}
\vspace{-0.2cm}
\label{tab:ablation_shanghai}
\begingroup\small
\resizebox{1\linewidth}{!}{
\begin{tabular}{c|cc}
\hline
\multirow{2}{*}{\textbf{Model}} & \multicolumn{2}{c}{\textbf{Shanghai}} \\
                             & Recall@10 & Recall@20 \\ \hline
PIGEON                       & 0.10503   & 0.14895      \\
w/o history record           & 0.02673   & 0.04838         \\
w/o personal history         & 0.07257   & 0.11776       \\
w/o similar users' history  & 0.06875   & 0.11585       \\
w/o need refinement          & 0.09039   & 0.14131     \\
w/o recall model fine-tuning  & 0.07257   & 0.10948     \\
w/o flexible needs in queries & 0.07731   & 0.11377 \\
w/o closed-set needs in queries & 0.09322   & 0.14438     \\
\hline
\end{tabular}}
\endgroup
\vspace{-0.2cm}
\end{table}

\vspace{-0.4cm}
\begin{figure}[h]
\centering
\subfigure[Beijing]{               
\includegraphics[width=3.5cm]{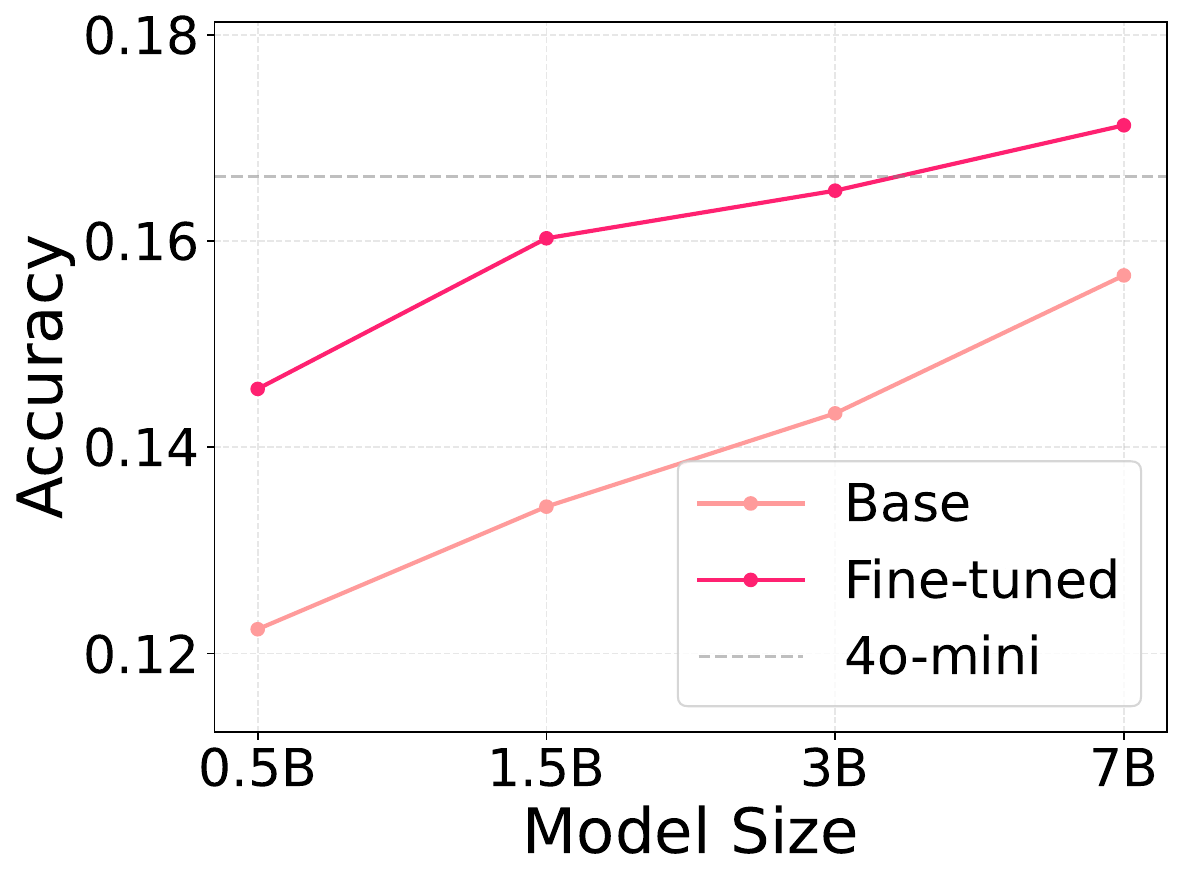}\label{fig:beijing_finetune}}
\hspace{0in}
\subfigure[Shanghai]{
\includegraphics[width=3.5cm]{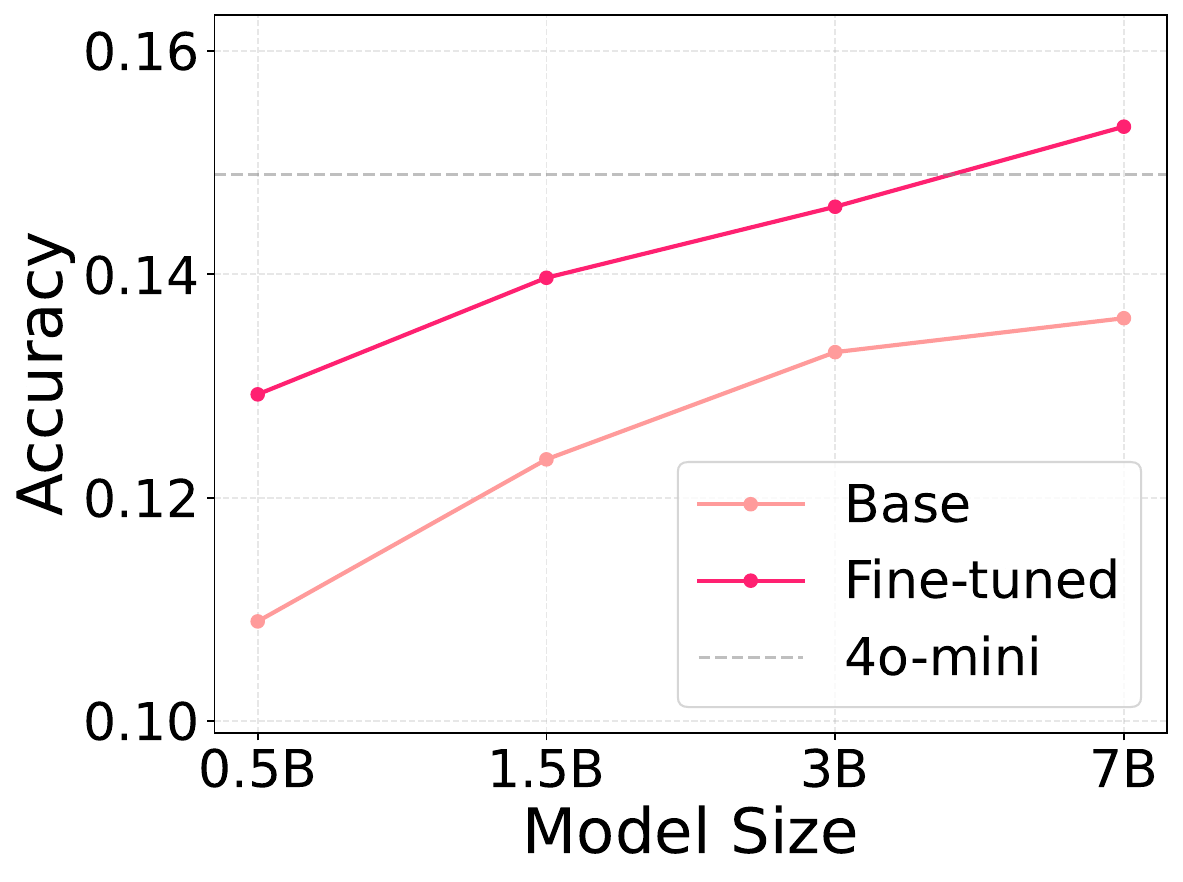}\label{fig:shanghai_finetune}}
\hspace{0in}
\vspace{-0.4cm}
\caption{Performance (Recall@20) of small LLMs on living need prediction after domain adaptation.}\label{fig:domain}
\vspace{-0.4cm}
\end{figure}

\subsection{Domain Adaptation of LLMs~(RQ4)}
To make our system practically deployable, we design an instruction tuning approach aimed at enabling smaller LLMs to achieve usable performance, thereby reducing computational costs and latency in online deployment. We conduct experiments using Qwen2.5 series models (0.5B, 1.5B, 3B, 7B) as base models, with results shown in Figure~\ref{fig:domain}. We find that through training, the models show significant performance improvements, with 3B models approaching the capabilities of GPT-4o mini, while the 7B model surpasses it. This demonstrates the effectiveness of our instruction tuning method. What's more, smaller LLMs can achieve competitive results with single-step inference, making our method more feasible for deployment.

\subsection{Performance on different LLMs (RQ5)}
\begin{figure}[]
\centering
\includegraphics[width=0.9\columnwidth]{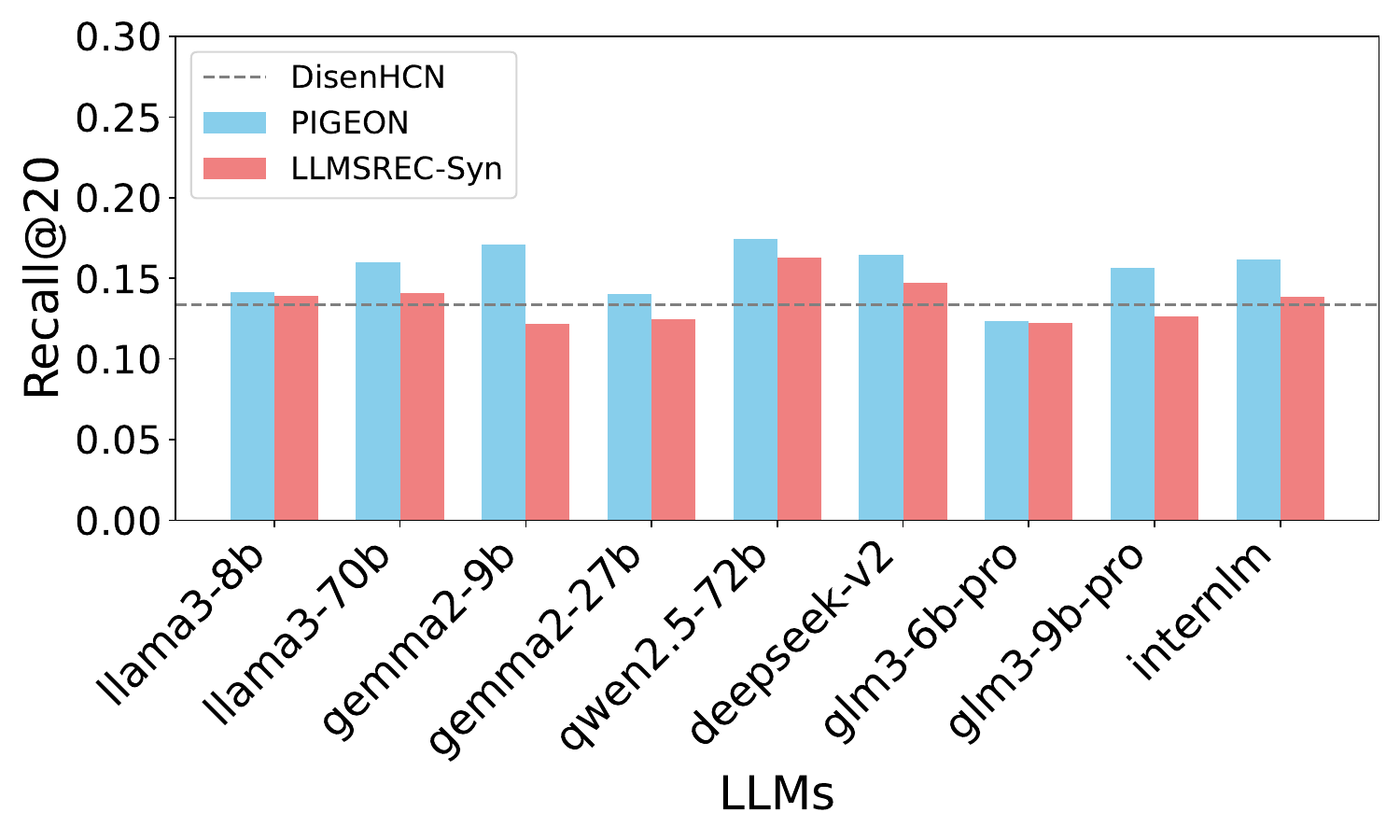}
\vspace{-0.2cm}
\caption{The performance of PIGEON and LLMSREC-Syn on various LLM backbones on the Beijing Dataset.} \label{fig:llms_bj}
\vspace{-0.35cm}
\end{figure}

\begin{figure}[]
\centering
\includegraphics[width=0.9\columnwidth]{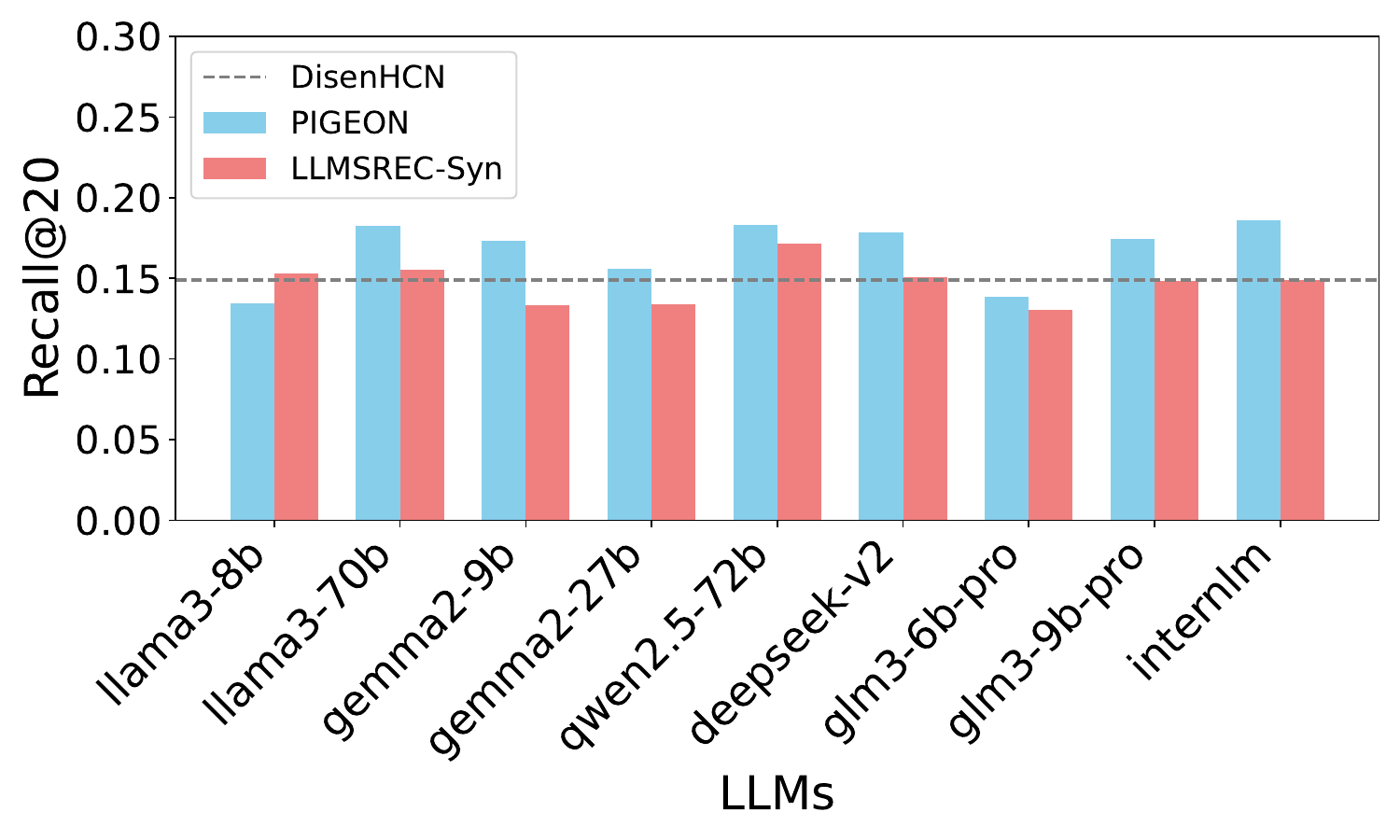}
\vspace{-0.2cm}
\caption{The performance of PIGEON and LLMSREC-Syn on various LLM backbones on the Shanghai dataset.} \label{fig:llms_sh}
\vspace{-0.35cm}
\end{figure}

In our experiments, we use GPT-4o mini as the main backbone of PIGEON. In this section, we test PIGEON's performance with several open-source LLMs to evaluate its effectiveness across these models. We also compare our method with LLMSREC-Syn, which uses a text embedding model to vectorize user and context for record retrieval. The results on the Beijing dataset are shown in Figure~\ref{fig:llms_bj} and~\ref{fig:llms_sh}, focusing on Recall@20.

The results indicate that even with smaller-scale open-source models, our method performs well, achieving performance comparable to or better than the best closed-set classification models. More importantly, our method outperforms the direct text embedding-based retrieval approach across all open-source models, demonstrating the effectiveness of our design.


\section{Related Work}
\subsection{Open-Set Classification}
Open-set classification extends traditional closed-set classification by handling scenarios where the model encounters unseen classes~\cite{scheirer2014probability}. Early methods used confidence thresholding to reject unfamiliar inputs~\cite{bendale2016towards}. Recent approaches incorporate generative models~\cite{mundt2019open} and adversarial learning~\cite{zhang2021deep, bao2022opental} for enhanced robustness. While recent advances in open-vocabulary object detection have shown promising results in handling unconstrained output spaces through vision-language models~\cite{zareian2021open,gu2021open,li2022language,pham2024lp}, extending beyond closed-set formulation remains unexplored for living need prediction. We address this by leveraging LLMs to generate flexible need descriptions without class constraints.
\subsection{Domain Applications of LLMs}
In recent years, LLMs have flourished and been applied to various domains. Successful application areas include natural sciences~\cite{li2025materials}, social sciences~\cite{gao2023s3,gao2024large,li2023econagent,piao2025agentsociety,wang2024survey}, urban science~\cite{feng2024citygpt,feng2024citybench, feng2025survey}, social media analysis~\cite{lan2024depression,lan2024stance}, reinforcement learning~\cite{hao2025rl, hao2025llm, xu2025towards}, literature analysis~\cite{hao2024hlm}, and graph learning~\cite{li2023survey,li2023gslb,zhang2024cut}. However, these methods primarily leverage the world knowledge of LLMs, while the flexible text generation capabilities of LLMs have not been fully utilized and may even become obstacles in structured tasks. In contrast, we effectively leverage both the commonsense and flexible text generation capabilities of LLMs to achieve better human need prediction.

\subsection{LLMs for User Modeling}
User modeling focuses on understanding users' behaviors and intentions to provide personalized experiences~\cite{feng2025agentmove}.LLMs have advanced this field through its commonsense knowledge, enabling success across tasks including profiling, recommendation, and behavior detection~\cite{wang2024can,liu2023llmrec}. However, no existing research has utilized LLMs for predicting user living needs, making our work pioneering in this domain.
\section{Conclusion and Future Work}\label{sec::conclusion}
In this study, we propose PIGEON, a novel open-set living need prediction system that leverages LLMs to generate flexible need descriptions unrestricted by pre-defined categories. Experimental results demonstrate that PIGEON outperforms both traditional closed-set and baseline LLM-based methods on need-based life service recall. Human evaluation validates the reasonableness and specificity of our generated need descriptions. Future work will focus on enhancing PIGEON's efficiency for large-scale industrial applications.
\clearpage

\section{Limitations}
While our approach demonstrates strong performance in open-set living need prediction, several limitations warrant discussion. First, the computational overhead of LLM inference operations poses scalability challenges for large-scale deployment, particularly in real-time scenarios requiring low latency. Second, our evaluation focused primarily on two cities within a single life service platform; broader validation across diverse geographical regions and service platforms would help establish the method's generalizability. Third, while our approach extends open-set classification beyond traditional boundaries, it may still be constrained by the LLM's ability to generate contextually appropriate need descriptions. Finally, the quality of predictions may vary depending on the choice of LLM backbone, as shown in our experiments with different models.

\section{Ethics Statement}
This work involves processing user behavioral data to predict living needs. We ensure all data is anonymized through multiple safeguards: user identifiers are replaced with random codes, locations are aggregated into several functional zones, and timestamps are discretized into 30-minute intervals. We further protect user privacy by removing personal identifiers from review text and filtering out unique behavioral sequences that could enable individual identification. No personally identifiable information is used in the study or retained in our datasets. 

The potential risks of behavioral data analysis are significant: malicious actors could potentially use predicted needs to manipulate user choices, create targeted advertising that exploits vulnerabilities, or combine predictions with other data sources to de-anonymize users. Commercial entities might attempt to leverage these predictions for aggressive marketing or price discrimination. There's also a risk of reinforcing existing biases in user behavior or creating filter bubbles that limit user exposure to diverse services. While we have implemented data obfuscation and anonymization to help reduce these risks, continuous ethical oversight remains necessary in practical applications.

The LLM components of our system utilize the GPT-4o mini API service provided by OpenAI, in full compliance with OpenAI's terms of service and usage policies. We maintain strict data handling protocols throughout our research to prevent potential privacy breaches or misuse.
\section{Acknowledgement}
This work was supported in part by the National Key Research and Development Program of China under grant 2024YFC3307603, in part by the China Postdoctoral Science Foundation under grant 2024M761670 and GZB20240384, in part by the Tsinghua University Shuimu Scholar Program under grant 2023SM235. This work was also supported by Meituan.

\bibliography{bibliography}
\clearpage

\appendix
\section{Appendix}
\subsection{The Structure of Maslow's Hierarchy Guided Living Needs Framework}
In our method, to ensure that predicted living needs align with actual user needs on life service platforms while maintaining flexibility, we integrate Maslow's hierarchy of needs with real-world platform services to create a structured framework. Below is the complete framework we developed, which guides our LLM in refining initial predictions to better reflect genuine human needs while preserving their diverse expressions:
\begin{styleditemize}
    \item Physiological Needs
    \vspace{-0.1cm}
    \begin{itemize}[topsep=0cm, itemsep=0cm, leftmargin=*]
        \item Basic Diet
        \begin{itemize}[topsep=0cm, itemsep=0cm, leftmargin=*]
            \item Home meals
            \item Restaurant dining
        \end{itemize}
        
        \item Housing Needs
        \begin{itemize}[topsep=0cm, itemsep=0cm, leftmargin=*]
            \item Basic living space
            \item Quality living environment
        \end{itemize}
        
        \item Daily Living Supplies
        \begin{itemize}[topsep=0cm, itemsep=0cm, leftmargin=*]
            \item Essential supplies
            \item Fresh groceries
        \end{itemize}
    \end{itemize}

    \item Safety Needs
        \vspace{-0.1cm}
    \begin{itemize}[topsep=0cm, itemsep=0cm, leftmargin=*]
        \item Health Protection
        \begin{itemize}[topsep=0cm, itemsep=0cm, leftmargin=*]
            \item General wellness maintenance
            \item Specialized medical care
            \item Preventive healthcare
        \end{itemize}
        
        \item Home Security
        \begin{itemize}[topsep=0cm, itemsep=0cm, leftmargin=*]
            \item Property protection
            \item Safety maintenance
        \end{itemize}
        
        \item Living Environment
        \begin{itemize}[topsep=0cm, itemsep=0cm, leftmargin=*]
            \item Cleaning services
            \item Environmental maintenance
        \end{itemize}
    \end{itemize}

    \item Social Belonging Needs
        \vspace{-0.1cm}
    \begin{itemize}[topsep=0cm, itemsep=0cm, leftmargin=*]
        \item Group Activities
        \begin{itemize}[topsep=0cm, itemsep=0cm, leftmargin=*]
            \item Dining and social venues
            \item Leisure and entertainment activities
        \end{itemize}
        
        \item Family and Parent-Child Interaction
        \begin{itemize}[topsep=0cm, itemsep=0cm, leftmargin=*]
            \item Parent-child activities and entertainment
            \item Family celebrations and event organization
        \end{itemize}
    \end{itemize}

    \item Esteem Needs
        \vspace{-0.1cm}
    \begin{itemize}[topsep=0cm, itemsep=0cm, leftmargin=*]
        \item Knowledge Development
        \begin{itemize}[topsep=0cm, itemsep=0cm, leftmargin=*]
            \item Educational training
            \item Arts and skill development
        \end{itemize}
        
        \item Career Growth
        \begin{itemize}[topsep=0cm, itemsep=0cm, leftmargin=*]
            \item Academic advancement
            \item Professional skills
        \end{itemize}
    \end{itemize}

    \item Self-Actualization Needs
        \vspace{-0.1cm}
    \begin{itemize}[topsep=0cm, itemsep=0cm, leftmargin=*]
        \item Cultural and Artistic Pursuits
        \begin{itemize}[topsep=0cm, itemsep=0cm, leftmargin=*]
            \item Cultural experiences and exhibitions
            \item Artistic creation and expression
        \end{itemize}
        
        \item Travel and Experiential Activities
        \begin{itemize}[topsep=0cm, itemsep=0cm, leftmargin=*]
            \item Travel and sightseeing
            \item Outdoor and experiential activities
        \end{itemize}
    \end{itemize}

\end{styleditemize}

\begin{figure*}[]
\centering
\subfigure[Impact of $K_p$, Beijing]{\includegraphics[width=0.46\columnwidth]{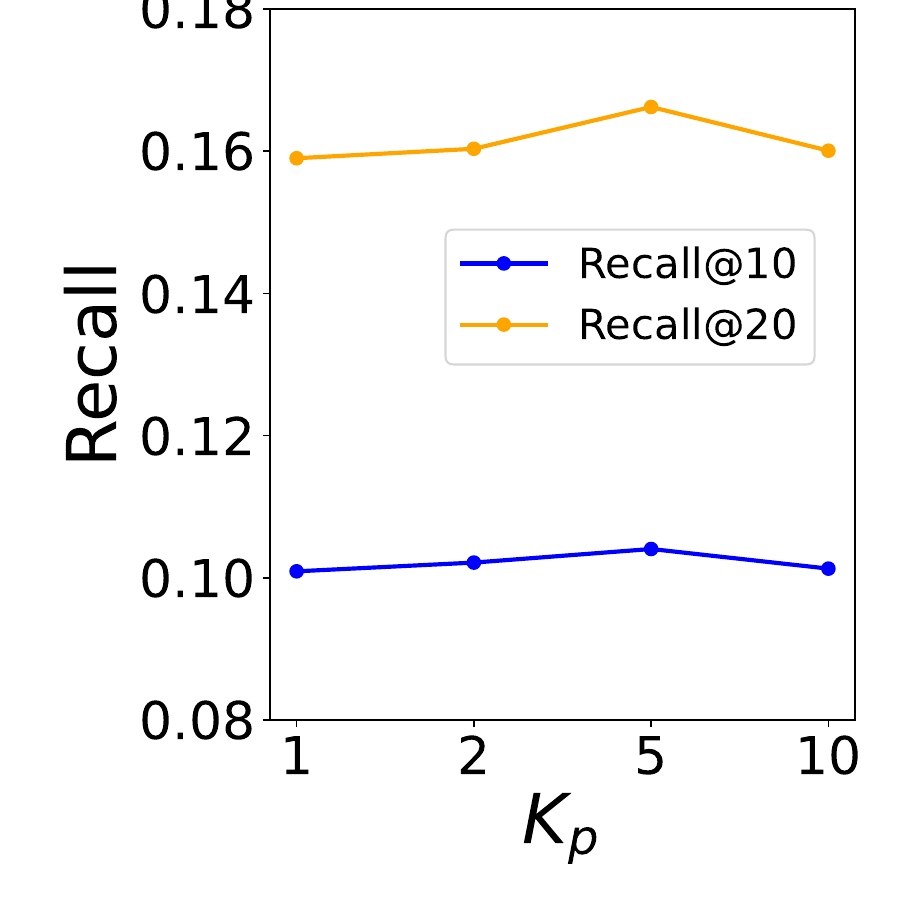}}
\subfigure[Impact of $K_s$, Beijing]{
\includegraphics[width=0.46\columnwidth]{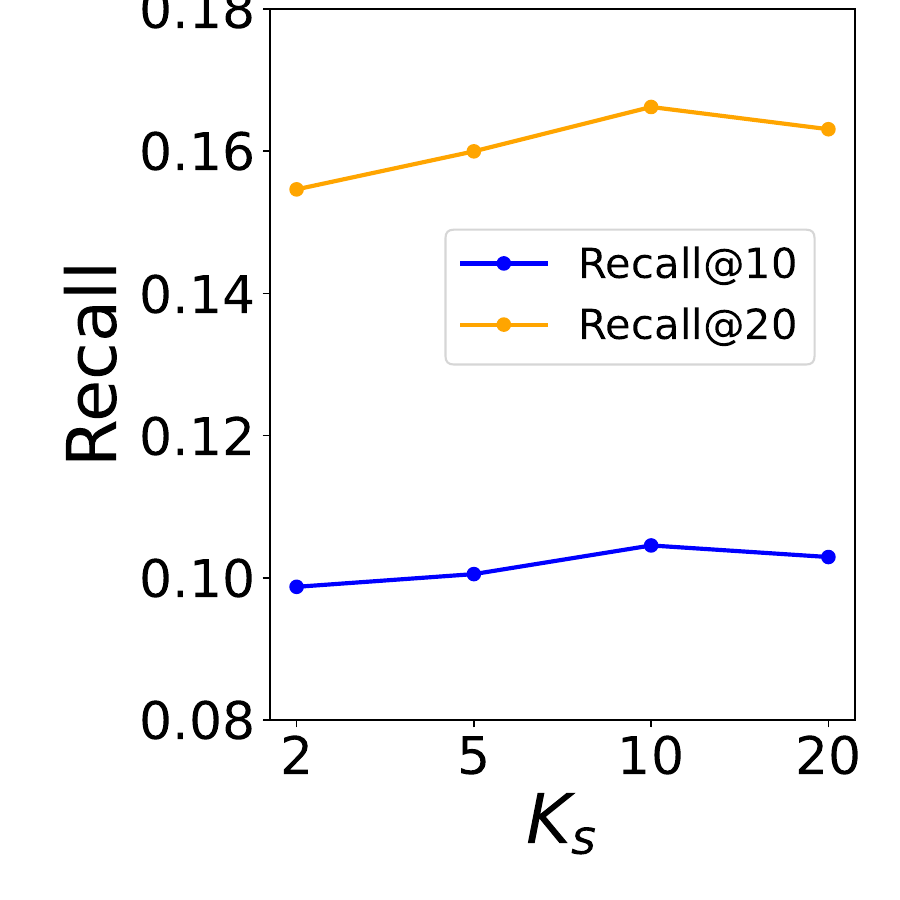}}
\subfigure[Impact of $K_p$, Shanghai]{
\includegraphics[width=0.46\columnwidth]{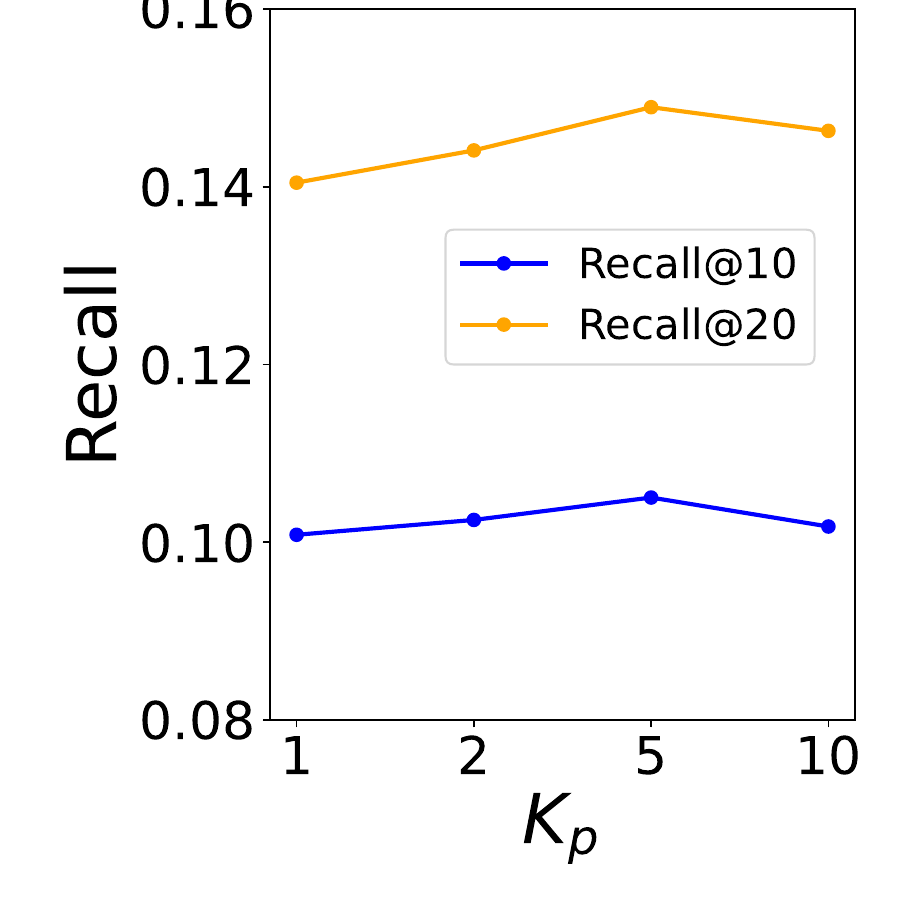}}
\subfigure[Impact of $K_s$, Shanghai]{
\includegraphics[width=0.46\columnwidth]{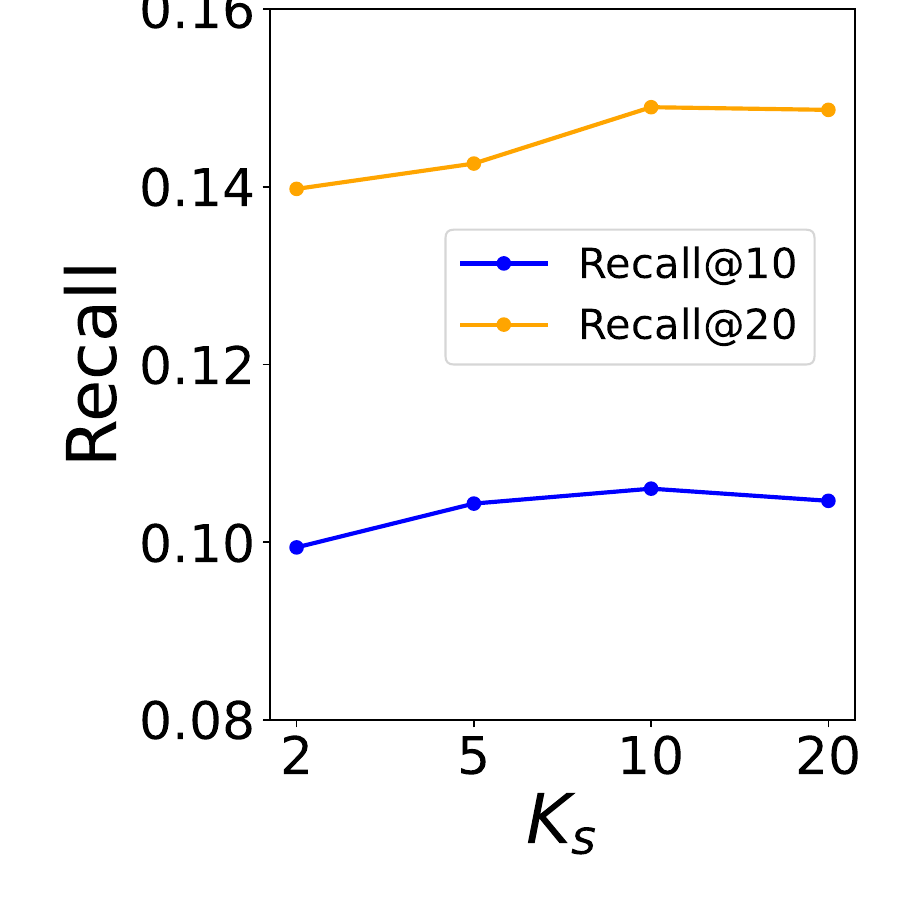}}
\caption{Results of hyperparameter study.} \label{fig:hyper}
\end{figure*}

\subsection{Hyperparameter Study~(RQ6)}
Our method introduces two hyperparameters: $K_p$, the number of personal relevant history records retrieved, and $K_s$, the number of relevant history records from similar users. Existing work~\cite{lin2024rella} reports that the quantity of historical records provided to the LLM as reference can influence prediction performance. To further investigate the finding in our task, we explore the impact of these two hyperparameters on our model's performance. The results are shown in Figure~\ref{fig:hyper}.

The results indicate that the optimal values for $K_s$ and $K_p$ are 5 and 10, respectively, on both datasets. Additionally, we observe that varying $K_s$ within [1, 2, 5, 10] and $K_p$ within [2, 5, 10, 20] has minimal effect on model performance. This demonstrates the stability of our method with respect to hyperparameter selection.

\begin{figure*}[t]
\centering
\includegraphics[width=16cm]{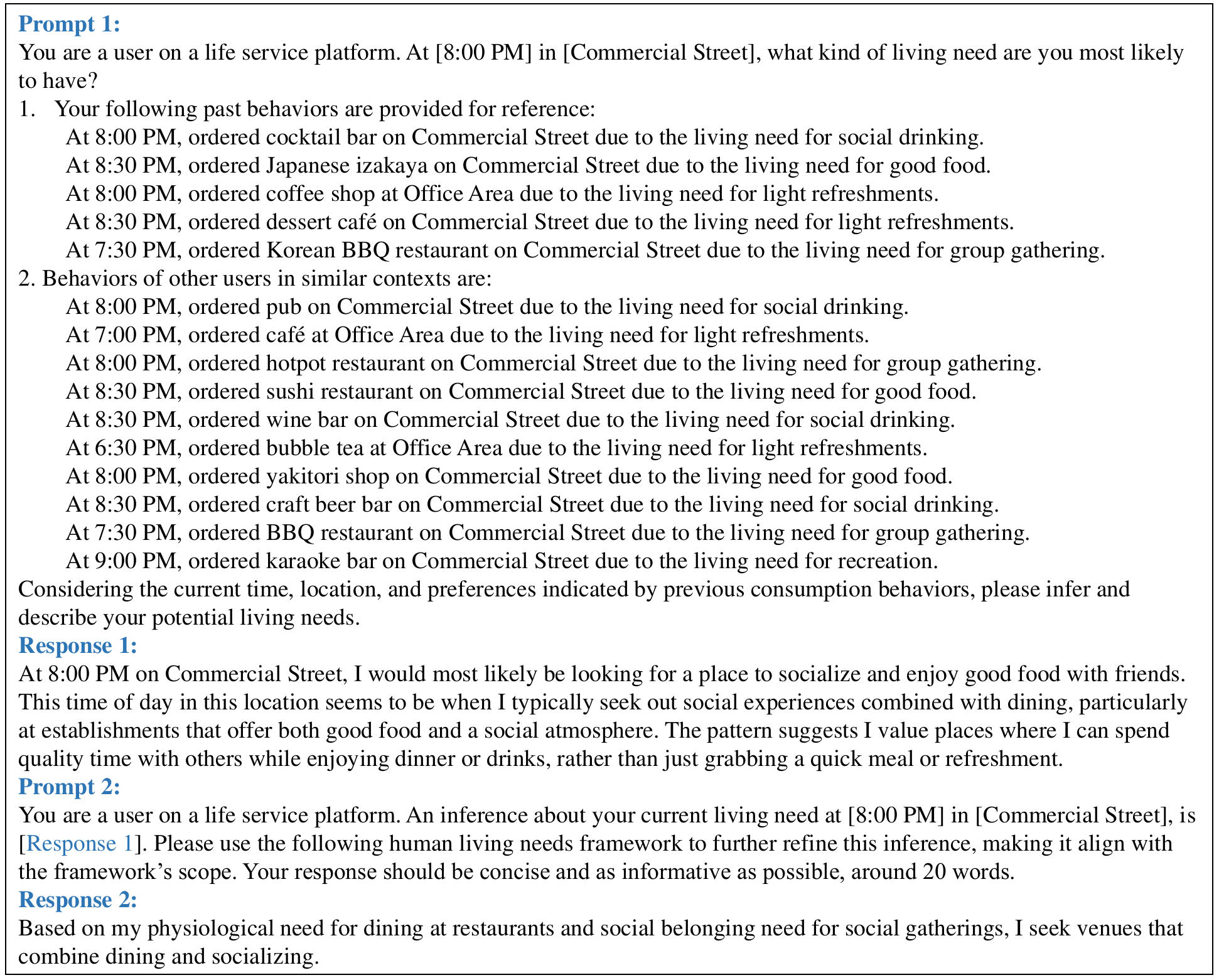}
\vspace{-0.2cm}
\caption{A working example of PIGEON.}
\vspace{-0.4cm}
\label{fig:workingexample}
\end{figure*}
\subsection{Hyperparameter Settings}
In our main method, the hyperparameters are set as follows:
\begin{itemize}
\setlength{\itemsep}{0pt}
\setlength{\parsep}{0pt}
\setlength{\parskip}{0pt}
    \item Number of personal historical records $K_p$: 5
    \item Number of similar users' historical records $K_s$: 10
    \item Number of GNN layers: $3$
    \item GNN embedding size: $64$
    \item GNN learning rate: $1\times10^{-4}$
    \item Triplet margin for triplet loss: $0.5$
    \item Learning rate for triplet loss: $2\times10^{-5}$
\end{itemize}
\subsection{Details of Human Evaluation~(RQ2)}
\label{sec:human}
The questionnaire uses a Likert scale format and consists of three groups of questions. Each group evaluates users' perceptions of need predictions from three models: PIGEON, LLMSREC-Syn, and DisenHCN. Each group contains 7 questions covering the five need categories from physiological to self-actualization needs, with two additional questions addressing common need combinations: physiological+social needs and physiological+safety needs. While PIGEON and LLMSREC-Syn generate predictions of approximately 10 words, DisenHCN provides fixed phrases. We evaluate three dimensions:
\begin{itemize}[leftmargin=*]
\setlength{\itemsep}{0pt}
\setlength{\parsep}{0pt}
\setlength{\parskip}{0pt}
\item Reasonableness: Assessing whether the predicted need is reasonable given the user's history and current context.
\item Specificity: Evaluating whether the prediction points to needs with concrete fulfillment methods and actionability.
\item Information Density: The amount of effective information conveyed per unit of text.
\end{itemize}
Each dimension is rated on a 5-point Likert scale (1=lowest, 5=highest). To prevent bias, participants are blinded to the source models and only presented with the predicted content.
To reduce cognitive load, we present only 5 behavioral records that demonstrate user preferences for each question. Here are sample questions from our survey:

(PIGEON)
\begin{center}
\begin{minipage}{0.92\linewidth}
\begin{shaded}
Time: 2:00 PM, February 8, 2024

Location: Commercial Street

Historical Records:
\begin{itemize}[leftmargin=*]
\setlength{\itemsep}{0pt}
\setlength{\parsep}{0pt}
\setlength{\parskip}{0pt}
\vspace{-0.3cm}
\item 3:00 PM, February 5, 2024: Hairdressing service at Commercial Street
\item 12:30 PM, February 1, 2024: Nail service at Commercial Street
\item 4:00 PM, January 28, 2024: Gym at Commercial Street
\item 11:00 AM, January 25, 2024: Yoga studio at Commercial Street
\item 2:30 PM, January 22, 2024: Eyelash service at Commercial Street
\end{itemize}
\vspace{-0.3cm}
Prediction: Esteem needs, seeking services for appearance enhancement and fitness maintenance to boost confidence.
\end{shaded}
\end{minipage}
\end{center}

(LLMSREC-Syn)
\begin{center}
\begin{minipage}{0.92\linewidth}
\begin{shaded}
[Same context and historical records as above]

Prediction: User likely seeking beauty or fitness services in the commercial street.
\end{shaded}
\end{minipage}
\end{center}

(DisenHCN)
\begin{center}
\begin{minipage}{0.92\linewidth}
\begin{shaded}
[Same context and historical records as above]

Prediction: Beauty
\end{shaded}
\end{minipage}
\end{center}

We recruit 116 participants, including graduate students in user modeling and business professionals from life service platforms. To ensure comprehension, we provide examples and detailed explanations of each dimension at the beginning of the questionnaire. To mitigate fatigue and order effects, we randomly select 15 questions from a pool of 21 for each participant to evaluate. The average completion time is approximately 11 minutes.

\subsection{Performance Optimization and Deployment Considerations}

Our method is designed for near-line deployment scenarios where second-level latency is acceptable. As described in Section 3.3, predicted needs serve as one source in the recall stage, and recall computations can be performed asynchronously. The computational flow is triggered by changes in time and location, recalculating only when time shifts to the next half-hour interval or location changes significantly, resulting in infrequent updates. While online computations often face strict latency constraints in the tens of milliseconds, near-line applications can tolerate latencies of seconds or even longer~\cite{li2021truncation}.

Despite moderate latency requirements, we have performed comprehensive optimizations to handle potential peak loads and ensure practical deployment feasibility.

\begin{table*}[]
\centering
\caption{Performance comparison of different inference optimization strategies on a single A100 GPU. P50/P99 refer to median and 99th percentile latency; QPS denotes queries per second.}
\label{tab:performance}
\begingroup\small
\resizebox{1\linewidth}{!}{
\begin{tabular}{c|c|ccc|ccc|ccc}
\hline
\multirow{2}{*}{\textbf{Method}} & \multirow{2}{*}{\textbf{Prefix Cache}} & \multicolumn{3}{c|}{\textbf{Batch Size = 64}} & \multicolumn{3}{c|}{\textbf{Batch Size = 128}} & \multicolumn{3}{c}{\textbf{Batch Size = 256}} \\
& & P50 (s) & P99 (s) & QPS & P50 (s) & P99 (s) & QPS & P50 (s) & P99 (s) & QPS \\ \hline
HF Transformers & - & 2.224 & 2.238 & 29.26 & 4.186 & 4.215 & 31.27 & 8.166 & 8.202 & 32.08 \\
vLLM & No & 0.298 & 0.501 & 210.80 & 0.427 & 0.441 & 303.60 & 0.706 & 0.881 & 354.57 \\
vLLM & Yes & 0.282 & 0.479 & 221.33 & 0.405 & 0.428 & 319.32 & 0.670 & 0.841 & 375.42 \\
SGLang & No & 0.260 & 0.435 & 232.96 & 0.362 & 0.374 & 329.31 & 0.629 & 0.758 & 415.43 \\
SGLang & Yes & 0.250 & 0.416 & 249.18 & 0.355 & 0.356 & 342.95 & 0.579 & 0.713 & 446.31 \\ \hline
\end{tabular}}
\endgroup
\vspace{-0.2cm}
\end{table*}

\subsubsection{LLM Inference Optimization}
We employ several strategies to reduce LLM inference latency:

\textbf{Model Fine-tuning:} We use instruction tuning on smaller LLMs. As shown in Figure~\ref{fig:domain}, a fine-tuned 3B parameter model achieves performance comparable to GPT-4o mini while significantly reducing computational requirements.

\textbf{Inference Engine Optimization:} We build upon existing frameworks like vLLM~\cite{kwon2023efficient} and SGLang~\cite{zheng2024sglang} that leverage techniques including KV Cache, PagedAttention, and RadixAttention for general LLM inference optimization.

\textbf{Targeted Prefill Optimization:} Since our average input length (approximately 400 tokens) is much larger than output length (approximately 20 tokens), the Prefill stage represents the main bottleneck. We restructure prompts to maximize common prefixes across queries without altering semantics. The optimized prompt structure is:

\begin{center}
\begin{minipage}{0.92\linewidth}
    \begin{shaded}
    \textit{You are a user on a life service platform. Given time and location, what kind of living need are you most likely to have? Considering the current time, location, and preferences indicated by previous consumption behaviors, please infer and describe your potential living needs. Answer with around 20 words.}
    
    \textit{Time: [time] Location: [location]}
    
    \textit{Your following past behaviors are provided for reference: [Personal Relevant Records]}
    
    \textit{Behaviors of other users in similar contexts are: [Similar Users' Relevant Records]}
    \end{shaded}
\end{minipage}
\end{center}

We implement acceleration by explicitly caching the common prefix (the first 60 tokens), similar to System Prompt optimizations in online chat services. We verify that this rewritten prompt maintains the model's predictive performance while enabling significant acceleration.

We test performance on a fine-tuned Qwen2.5-3B model running on a single A100 GPU. Results are shown in Table~\ref{tab:performance}, demonstrating significant latency reduction while maintaining accuracy. Our optimization achieves a 13.9× acceleration compared to Hugging Face Transformers inference.

Using SGLang with Prefix Caching at batch size 256, our method maintains P99 latency below 0.75 seconds while processing over 440 queries per second. There exists a trade-off between latency and QPS: larger batch sizes allow more queries to be processed in batches, increasing QPS but also increasing latency per query; smaller batch sizes reduce latency per query but decrease QPS. In actual deployment, batch size should be dynamically adjusted based on real-time load, potentially leveraging queuing theory. For latency-critical scenarios, setting batch size to 1 achieves an average latency of 157ms.

Crucially, the optimizations we implement are strictly lossless, incurring no loss in accuracy. In practical applications, further acceleration can be achieved by incorporating techniques such as quantization and FP8 mixed-precision inference. Simple calculations indicate that using at most 8 A100 80G GPUs, our system can handle over 10 million queries per hour with no accuracy loss, demonstrating that online deployment is entirely feasible considering the inherently infrequent computation requirements of this method.

\subsubsection{Other Component Optimizations}
\textbf{Behavioral Embedding Learning and Retrieval:} User and context embeddings can be pre-computed offline, incurring no additional inference latency. Vector similarity search leverages mature open-source solutions like Milvus~\cite{wang2021milvus}, which achieves throughputs of 15,000-20,000 QPS for nearest neighbor search on millions of vectors using cost-effective hardware while maintaining recall rates above 0.95. We plan to deploy using such established solutions, with potential targeted optimizations to further enhance performance.

\textbf{Life Service Recall:} The recall process, mapping flexible need descriptions to services, can also achieve low latency. Experiments on a consumer-grade GPU (Nvidia 4090) show that vectorizing 10,000 predicted need descriptions (output limited to approximately 20 words by prompt constraints) using the fine-tuned bge-base-en-v1.5 model takes an average of 8.92ms per description. Service vectors can be pre-computed, and matching need embeddings to the nearest service vectors can be efficiently handled by systems like Milvus, typically completing within 10ms.

These comprehensive optimizations demonstrate that online deployment is entirely feasible while maintaining the system's prediction accuracy and meeting practical performance requirements for large-scale industrial applications.

\subsection{Details of Comparison Methods}
\label{sec:baselines}
Here, we give a more detailed discussion of our utilized comparison methods.

Closed-set prediction methods:
\begin{itemize}[leftmargin=*]
\item \textbf{XGBoost}~\cite{chen2016xgboost}: A widely used gradient-boosting model that builds an ensemble of decision trees to capture nonlinear interactions between user, time, and location features. Known for its robustness and strong generalization capabilities, XGBoost is effective in CTR tasks where feature relationships are complex and varied.
\item \textbf{DeepFM}~\cite{guo2017deepfm}: This model integrates a deep neural network with a factorization machine (FM) to learn both low- and high-order feature interactions, such as the cross-features between user, time, and location. DeepFM effectively captures intricate dependencies without requiring extensive feature engineering, making it a strong baseline for closed-set need prediction.
\item \textbf{DCN}~\cite{wang2017deep}: The Deep \& Cross Network (DCN) combines a deep network and a cross network, which explicitly models cross-feature interactions in a multi-layered structure. By learning high-level feature combinations, DCN can model complex dependencies among user, time, and location, increasing the prediction accuracy within fixed need sets.
\item \textbf{XDeepFM}~\cite{lian2018xdeepfm}: Extending DeepFM, XDeepFM includes a Compressed Interaction Network (CIN) to capture complex, higher-order feature interactions across layers. This component enhances the model’s ability to learn subtle cross-feature patterns between user, time, and location, making it a powerful model for structured need prediction tasks.
\item \textbf{EulerNet}~\cite{tian2023eulernet}: A recent CTR model optimized for large-scale recommendation, EulerNet efficiently handles high-dimensional feature inputs. It incorporates a multi-layered structure and leverages feature embeddings to support large-scale applications with diverse features, providing competitive performance in need prediction tasks.
\item \textbf{LightGCN}~\cite{he2020lightgcn}: An effective and concise Graph Convolutional Network (GCN) model. In our implementation, LightGCN constructs graphs with user, spatiotemporal context, living needs, and life services as nodes. The graph includes four types of edges: user-living need edges, user-life service edges, spatiotemporal context-living need edges, and spatiotemporal context-life service edges. The training method for LightGCN follows the same approach as the encoder training described in the paper. By focusing on core graph embeddings and omitting complex transformation layers, LightGCN efficiently captures direct relationships, making it effective for need prediction based on user and context.
\item \textbf{HAN}~\cite{wang2019heterogeneous}: The Heterogeneous Attention Network (HAN) constructs a multi-relational graph, incorporating diverse types of nodes and edges (e.g., user, time, location, need) into a unified structure. HAN uses attention mechanisms to weigh different types of relationships, allowing it to learn the relative importance of each edge in predicting closed-set needs within varied contexts. In our implementation, we construct meta-paths including user-need-user, user-service-user, and context-need-context to capture different semantic relationships, then apply node-level and semantic-level attention to learn comprehensive representations.
\item \textbf{DisenHCN}~\cite{li2022disenhcn}: DisenHCN utilizes disentangled representations within a hypergraph framework, breaking down user preferences along dimensions like time and location. By isolating preference factors in separate subspaces, DisenHCN captures the unique aspects of user behavior in each dimension, providing a more refined prediction of user needs.
\end{itemize}
Open-set prediction methods:
\begin{itemize}[leftmargin=*]
\item \textbf{Zero-Shot CoT}~\cite{kojima2022large}: The Chain-of-Thought (CoT) zero-shot approach leverages the LLM’s generalization capability to predict needs based solely on current time and location, bypassing the need for historical data. This method serves as a baseline to evaluate how well the model can deduce user needs purely from situational context.
\item \textbf{ReLLa}~\cite{lin2024rella}: ReLLa enhances LLM-based predictions by employing a text embedding model to convert user behavior history and spatiotemporal context into a vectorized format. During prediction, ReLLa retrieves the most relevant historical user records based on the current time and location, improving need prediction accuracy by leveraging similar past behaviors.
\item \textbf{LLMSRec-Syn}~\cite{wang2024whole}: Extending ReLLa, LLMSRec-Syn retrieves similar historical behaviors not only from the target user but also from other users with similar patterns. This method leverages both individual and similar users' behavior, enhancing the open-set need prediction process by incorporating both user-specific and community-based context.
\end{itemize}

\subsection{A Working Example of PIGEON}

Here, we provide a complete working example of PIGEON, as shown in Figure~\ref{fig:workingexample}. It can be seen that first, a comprehensive and multi-faceted need description was generated. Then, PIGEON generates a more concentrated and structured need expression. This refined need expression can be used for downstream applications such as life service recall.

\subsection{Rationale for Using Maslow's Hierarchy}\label{sec:maslow}
The choice of Maslow's hierarchy as our theoretical foundation is motivated by several key considerations. First, Maslow's model provides a comprehensive yet structured framework that naturally aligns with the hierarchical organization of life services on platforms. For instance, food delivery services correspond to physiological needs, while educational services map to higher-level growth needs. This natural alignment facilitates more systematic refinement of LLM predictions.

Second, Maslow's theory has been extensively validated in consumer behavior research~\cite{subrahmanyan2008integrated,schewe1988marketing,cao2013maslow}, demonstrating strong correlations between its need levels and consumer purchase patterns. This empirical foundation makes it particularly relevant for our task of predicting needs that drive service consumption, as it helps ensure our predictions align with established patterns of human motivation and consumption behavior.

While simpler prompting approaches could potentially guide the LLM to avoid overly specific predictions, Maslow's framework offers additional benefits by providing a structured way to organize and validate need predictions against established psychological principles. Our ablation studies show modest but consistent improvements with this approach, suggesting it helps maintain prediction quality while adding theoretical grounding.
\subsection{Cost Analysis}
In our main experiment, we used GPT-4o-mini as the base large language model. The costs per 10,000 predictions are approximately:
\begin{itemize}
\item Zero-shot CoT: \$0.09
\item ReLLa: \$0.21
\item LLMSREC-Syn: \$0.67
\item PIGEON: \$1.05
\end{itemize}
\newpage 
\subsection{The List of Closed-Set Living Needs}
For traditional closed-set living needs prediction, we extracted needs from user reviews and categorized them into 21 types under expert guidance. These living needs are:

\begin{styledenumerate}
    \item Good food
    \item A quick meal
    \item Cooking at home
    \item A group meal
    \item Social drinking
    \item Business travel
    \item Personal grooming
    \item Group gathering
    \item Home supplies
    \item Physical fitness
    \item Pet companionship
    \item Light refreshments
    \item Family bonding
    \item Vehicle service
    \item Recreation
    \item Body relaxation
    \item Beauty
    \item Leisure travel
    \item Child education
    \item Skill development
    \item Daily essentials
\end{styledenumerate}

\end{document}